\def\year{2020}\relax
\documentclass[letterpaper]{article} 
\usepackage{aaai20}  
\usepackage{times}  
\usepackage{helvet} 
\usepackage{courier}  
\usepackage[hyphens]{url}  
\usepackage{graphicx} 
\def\UrlFont{\rm}  
\usepackage{graphicx}  
\usepackage[utf8]{inputenc} 
\usepackage[T1]{fontenc}    
\usepackage{url}            
\usepackage{booktabs}       
\usepackage{amsmath,amssymb,amsfonts}
\usepackage{nicefrac}       
\usepackage{microtype}      
\usepackage{cite}
\usepackage{graphicx}
\usepackage{xcolor}
\usepackage{todonotes}
\usepackage{wrapfig}
\usepackage{enumitem}
\usepackage[font=small,labelfont=bf, skip=20pt]{caption}

\DeclareMathOperator*{\minimize}{minimize}
\DeclareMathOperator*{\argmax}{argmax}

\title{Exploring Adversarial Examples via Invertible Neural Networks}
\author{%
  Ruqi Bai \hspace{3mm}
  Saurabh Bagchi \hspace{3mm}
  David I. Inouye \\ \\
  School of Electrical and Computer Engineering \\
  Purdue University \\
  \texttt{\{bairuqi,sbagchi,dinouye\}@purdue.edu}
  }
\begin{document}

\maketitle

\begin{abstract}
Adversarial examples (AEs) are images that can mislead deep neural network (DNN) classifiers via introducing slight perturbations into original images.
This security vulnerability has led to vast research in recent years because it can introduce real-world threats into systems that rely on neural networks.
Yet, a deep understanding of the characteristics of adversarial examples has remained elusive.
We propose a new way of achieving such understanding through a recent development, namely, invertible neural models with Lipschitz continuous mapping functions from the input to the output.
With the ability to invert any latent representation back to its corresponding input image, we can investigate adversarial examples at a deeper level and disentangle the adversarial example's latent representation.
Given this new perspective, we propose a fast latent space adversarial example generation method that could accelerate adversarial training.
Moreover, this new perspective could contribute to new ways of adversarial example detection.
\end{abstract}

\section{Introduction}
\label{sec:intro}
Image classification problems have achieved great success using deep neural networks (DNNs).
However, recent work has reported that DNN classifiers have a security issue such that small perturbations to the input that humans may not recognize can change the result of the classifier \cite{Goodfellow2015, Kurakin2016a, Papernot_SP16}.
This security vulnerability is critical, since it means that for example, an adversary can make an autonomous vehicle mis-recognize a stop sign as an yield sign \cite{Papernot2017}. 
There has been a tremendous amount of work in generating and defending against adversarial examples, leading to a veritable arms race in the security literature. {\bf However, still elusive are answers to two fundamental questions}: (i) What are the precise characteristics of the examples produced by the adversarial example generators (called Adversarial Examples or AEs)? and (ii) What is fundamental in deep models that make them vulnerable to AEs? 

In this paper, we introduce a new perspective to begin the journey to answer these two questions --- we explore adversarial examples via invertible neural networks. Invertibility of the neural network allows us to get a better understanding the relationship between the input and the output by manipulating the output or middle layer and inverting back to the input space to investigate the difference in the input space between legitimate and adversaral images.
\begin{figure*}[!htbp]
    \centering
    \includegraphics[width=0.9\linewidth]{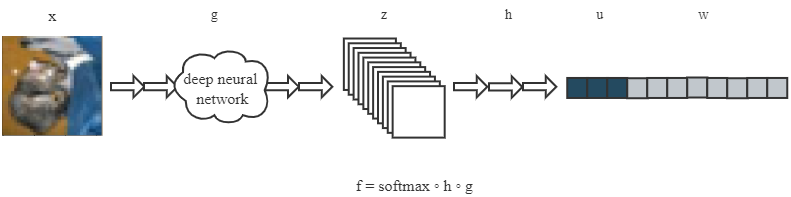}
    \vspace{-2em}
    \caption{A conceptual view of our reasoning framework for understanding Adversarial Examples (AEs) and notation in the paper. We use $x$ as the input image, where $x\in\mathcal{X}$, $z=g(x), z\in\mathcal{Z}$ is the output of the residual blocks, which are represented by $g$. We denote the output of the classifier $h$ as $(u,w) = (h_1(z), h_2(z))=h(z)$, where $u$ represents the logits and $w$ represents the non-logits.  The classification probabilities $p$ are merely softmax applied to the logits $u$, i.e., $p=softmax(u)$. We denote the full model as $f=softmax\circ h_1\circ g$, so $p=f(x)$. $x'$ denotes the adversarial examples with corresponding $(u', w')$.}
    \label{fig:notation}
\end{figure*}

{\em Our investigation reveals two insights relevant to the two questions above.} {\em First}, we find that AEs not only flip the logits (which is expected since they aim to change the result of the classification) but also change the non-logits (which was not expected).
{\em Second}, we hint at a fast method to generate adversarial examples. By creating a diverse set of AEs and fast, this can aid in making deep models robust through adversarial training. This is promising since adversarial training has emerged as one of the few promising approaches to defend against adversarial examples \cite{pgd_attack}. However, its success has been limited by the fact that generating high quality adversarial examples, with the Carlini-Wagner approach \cite{CW_attack} defining the gold standard, has been excruciatingly slow because it solves a multi-term optimization problem in a high-dimensional space.

\section{Background}
\subsection{Threat Model}
To understand the characteristics of AEs, we consider three representative AE generation techniques: Fast Gradient Sign Method (FGSM) \cite{Goodfellow2015}, Projected Gradient Descent(PGD) \cite{pgd_attack} and Carlini \& Wagner(CW) \cite{CW_attack}. We chose these because FGSM is the fastest way, PGD is widely used as a good balance between speed and quality of AE, and CW is the state-of-the-art attack technique which has been successful against all defenses available today, produces small perturbation in the input space, but is very slow.
FGSM  perturbs all input pixels by the same quantity $\epsilon$ in the direction of a gradient sign, \textit{i.e.}, $x' = x + \epsilon \text{ } \text{sign}\left(\frac{\partial J(x,y)}{\partial x} \right)$
where sign$(v)$ is 1 if $v>0$, -1 if $v<0$, and 0 if $v=0$.
In other words, FGSM attempts to make an AE by adding a noise to each pixel in the direction that maximizes the increment in the cost function.
The value of $\epsilon$ is chosen to be a multiple of $\epsilon_0$, which corresponds to the magnitude of one-bit change in a pixel. 
PGD iteratively applies FGSM (say, $N$ times) with the minimal amount of perturbation at a time with a random start point $x_0$ within the $\ell$ norm ball around $x$ as follows: $x_{n+1} = \text{Clip}_{X, \epsilon}\left\{ x_{n} + \alpha \textnormal{ sign}\left(\frac{\partial J(x_{n},y)}{\partial x_{n}} \right)\right\}$
with $x_0=x+\alpha$ and $x_N=x'$. Here, Clip$\{\cdot\}$ denotes a pixel-wise clipping operation, which ensures that the pixel value stays in the $\epsilon$-vicinity of the original value, and in the valid range. 
\emph{Carlini \& Wagner} (CW) minimizes a loss function containing two parts: the first part is the perturbation level, which is usually an $L_p$ norm of $\delta$ while 
the second part contains the term that tries to flip the prediction, i.e.,
$\minimize_{\delta} \|\delta\|_p + c \cdot \ell(x+\delta) \,$
where $\ell$ is a specially designed function (see \cite{CW_attack} for exact forms) to encourage label flipping such that $\ell(x+\delta)<0$ only if $L(x')\neq L(x)$. 
 
\subsection{Invertible Residual Network Structure}

The Invertible Residual Networks (i-ResNets) \cite{invertible} ensures the invertibility of a residual block by controlling the Lipschitz constant of the residual part.
In particular, i-ResNets ensures that the residual part has a Lipschitz constant below 1 by normalizing the linear operations (including convolutions).
This ensures that each block is invertible and can be inverted using a fixed point method.
Compared to other invertible networks such as those used in normalizing flows \cite{realnvp, glow}, this construction is quite flexible and does not require as many constraints on the architecture. The authors also show that these networks are bi-Lipschitz (i.e., Lipschitz bounded in both the forward and inverse directions) and thus may be more stable than previous invertible networks.
Finally, \cite{invertible} showed that these networks can achieve comparable classification performance despite being invertible (for the evaluated datasets CIFAR-10, CIFAR-100, and MNIST).
Because of these benefits, we choose i-ResNets for our exploration of adversarial examples.

\begin{figure}
\vspace{-3em}
  \begin{center}
      \includegraphics[width=0.9\linewidth]{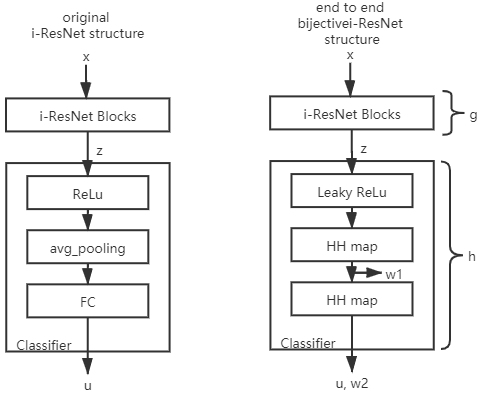}
      \vspace{-2em}
    \caption{
    We modify the i-ResNets for classification to be end-to-end bijective by changing ReLUs to leaky ReLUs and linear maps to Householder reflections.
    Note that $w1$, $w2$ represent the two parts of non-logits $w$; and HH means Householder reflection}
    \label{fig:structure}
    \vspace{-2em}
\end{center}

\end{figure}

\section{Our Exploration of Adversarial Examples}\label{exps}
\begin{figure*}[!htbp]
    \centering
    \includegraphics[width=0.9\linewidth]{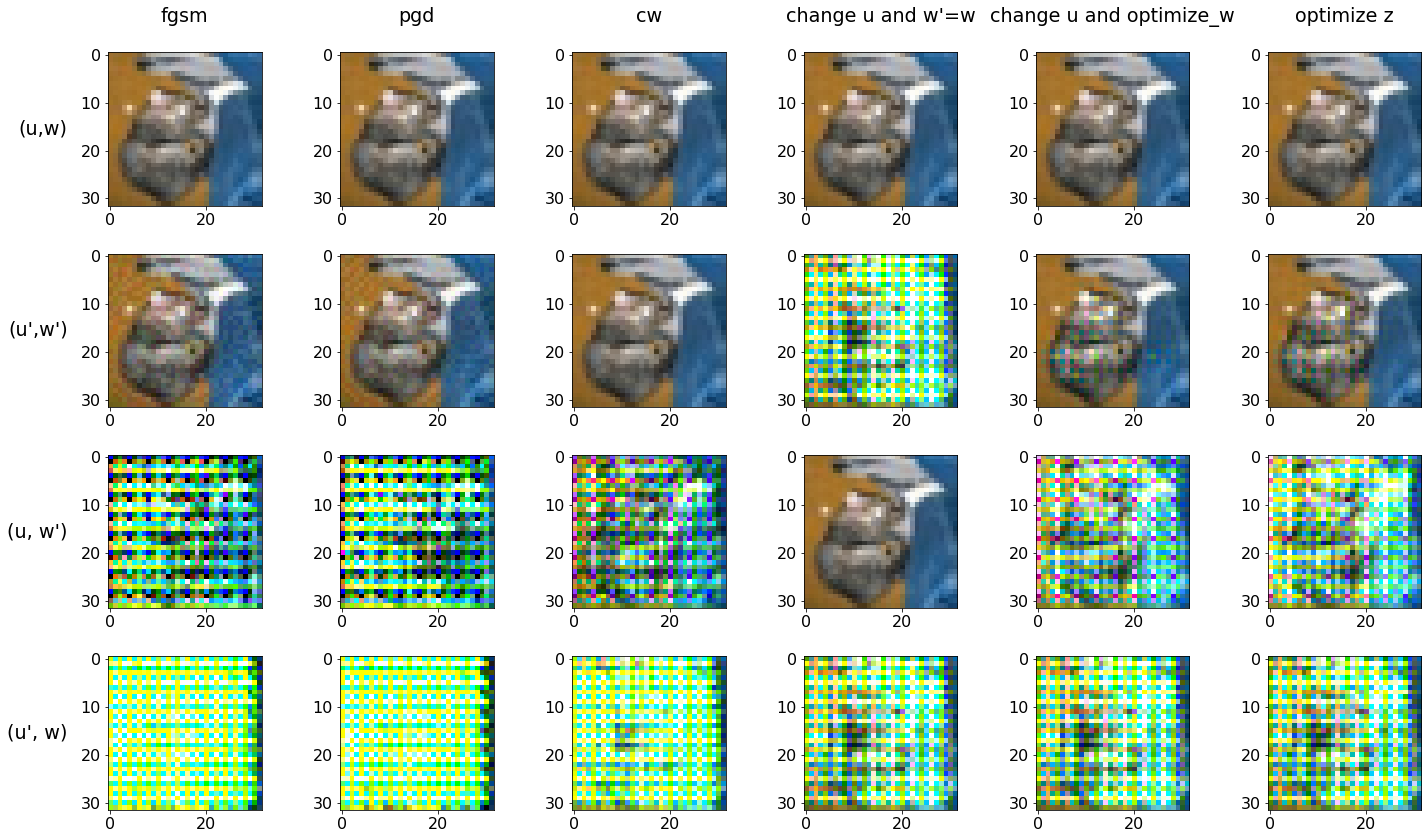}
    \vspace{-1.5em}
    \caption{These examples of disentangling the logits $u$ and the non-logits $w$ show that prior AE methods (FGSM, PGD, CW in first 3 columns) must modify both $u$ and $w$ since modifying one alone does not result in an AE (bottom two rows).
    We propose three novel methods (right three columns) for AE generation by optimizing $u$ and $w$ directly rather than optimizing the input image $x$.
    We see that only modifying the logits results in large changes in the input (4th column, 2rd row) while if we modify the logits $u$ and then optimize the non-logits $w$(5th column), we can arrive at a reasonable AE in a fraction of the time. We could also optimize $u$ and $w$ simultaneously which achieve better performance.
    The rows show in order: the benign example $(u,w)$, the AE $(u', w')$, the example with benign logits but adversarial non-logits $(u,w')$, and the example with adversarial logits but benign non-logits $(u', w)$. More examples shows in the Appendix \ref{sec:moreexp}
    }
    \label{fig:all}
    \vspace{-3em}
\end{figure*}
\begin{table*}[!hbp]
\vspace{-5em}
\caption{The performance and perturbation of different attacks on Invertible Residual Network show that the perturbation in the input space is related to the perturbation in the latent $\mathcal{Z}$ space.
Also, we note that our attack by optimizing $w$ creates a misclassification by construction of the logits (thus 0\% accuracy), has a very small latent perturbation, and is much faster than PGD or C\&W but our method induces a somewhat larger input perturbation. The time is that for generating a single AE. This shows the extreme slowness of C\&W, while our approaches are 4X and 6.5X faster than PGD, the feasible approach used today for adversarial training.
}
\label{tab: attack}
\centering
\begin{tabular}{@{}lllllll@{}}
\toprule
Attack & Accuracy & $\lVert x-x'\rVert_2$ & $\lVert x-x'\rVert_\infty$ & $\lVert z-z'\rVert_2$    & $\lVert z-z'\rVert_\infty$ & Time(in s) \\ \midrule
FGSM          & 20.98\%                     & 1.7326          & 0.0314            & 25.3375 & 5.4419 & 0.0081         \\
PGD           & 7.71\%                      & 1.5171          & 0.0314            & 23.9304 & 4.9491 & 0.3185         \\
C\&W          & 7.81\%                    & 0.0851          & 0.0135            & 1.9037  & 0.5702 &  44.741          \\ 
$w$ attack (Ours)      & 0.00\%                    & 9.5729          & 0.8805            & 0.0593  & 0.0165 &      0.0778         \\
$z$ attack (Ours)      & 18.66\%                    & 3.2349          & 0.4829            & 0.0348  & 0.0103 &      0.0489         \\\bottomrule
\end{tabular}
\end{table*}
Most classifiers map a high-dimensional input image to a low dimensional output, i.e., the classification logits $u$. 
A majority of information is lost during this mapping~\cite{bronevetsky2012automatic}.
However, with our bijective network, we explicitly retain all the classification irrelevant information in the non-logits $w$.
Thus, we can disentangle how current adversarial methods modify the logits as well as the non-logits.
First, we explore the three standard adversarial attacks on our invertible neural network and discover that AEs significantly modify both the logits {\em and} the non-logits simultaneously.
This is somewhat surprising since the goal of AE attacks is only to flip the label, i.e., modify the logits, but our results suggest that the current attack methods indeed also modify even the irrelevant latent representation (i.e., the non-logits).
Next, we propose a novel method to generate AEs by first modifying the latent representation and then using the invertibility property of our network to obtain adversarial images, i.e., AEs in the input space. 
Our results show that it is possible to efficiently attack only the last classifier layer and corresponding latent representation to produce high quality AEs, which opens up the door for effective adversarial training. 

\paragraph{Preparation: End-to-end bijective}

To utilize the invertible residual network in the adversarial scenario, we first need the model to be end-to-end bijective.
The original i-ResNets code only guarantees invertibility in the residual block parts. 
The classifier block of the i-ResNets classifier network contains a ReLU function, an average pooling, and a linear transformation (See Figure \ref{fig:structure}), which are all not bijective.
Hence, we modify the last layer to make it fully bijective as in Figure \ref{fig:structure} using leaky ReLUs and Householder transformations in place of the usual blocks (further details are in the Appendix).
Our fully bijective network has a top-1 accuracy of $90.00\%$.

\subsection{Exploring Current Adversarial Attacks}

This section focuses on understanding three current popular attack methods on our bijective i-ResNets.
To do this, we separate the attacked sample $x'$ into its logits $u'$ and non-logits $w'$ components in the classifier block and replace either $u'$ or $w'$ with the original sample's (i.e., $x$'s) $u$ or $w$.
We want to understand whether current attack methods modify just the logits (i.e., the part important to classification) or modify both the logits and the non-logits simultaneously.
Figure \ref{fig:all} shows the results of attacks on end-to-end i-ResNets.
As one key observation, we notice that either adversarial change by itself (i.e., to either $u$ or $w$) produces out of distribution images; only when manipulating both the logits and the non-logits are we able to form valid AEs.
Table \ref{tab: attack} shows the model accuracy on AEs and also the perturbation of the adversarial examples in both $X$ and $Z$ space (after normalization). 
C\&W comparing to FGSM and PGD, has much smaller perturbation both in input space and latent space.

\subsection{Adversarial Attack towards the last layer}
Since i-ResNets is bi-Lipschitz, we have that $\lVert x'-x \rVert\leq L\lVert z'-z \rVert$ for Lipschitz constant $L$, i.e., the distance between AEs and benign examples in $\mathcal{X}$ space can be controlled by the distance of their latent representations in $\mathcal{Z}$ space. 
Thus, we propose to directly attack $\mathcal{Z}$ space instead of $\mathcal{X}$ space to avoid heavy computation of the residual neural network $g$ and only work with the last invertible classification layer $h$.
To generate effective AEs, we need to manipulate the logits to ensure classification flips, i.e., $argmax(u')\neq argmax(u)$, but also try to minimize the distance in $\mathcal{Z}$ space so that the example will only change $z$ slightly.
We discuss both of these steps below.



By directly changing the original logits, we can ensure that the example is misclassified by construction.
While there are many ways to manipulate the logits to get a misclassification, we choose a simple way which seems to be effective in our experiments:
    (1) Choose the largest probability value and the target class probability value (in untargeted attack, the target class is the one with the second largest probability).
    (2) Calculate the average of the two probability values $\bar{p}$.
    (3) Assign the target class a probability of $\bar{p}+\epsilon$ and the original class a probability of $\bar{p}-\epsilon$.
Given our logits $u'$ (which are misclassified by construction), we are free to modify the non-logits $w$ in any way to make the perturbation small.
The most intuitive idea is that we do not modify the non-logits at all from the original image.
However, we have seen experimentally that this method does not generate valid AEs (Figure~\ref{fig:all}, row $(u', w)$).
Small changes of logits can create large shifts in $\mathcal{Z}$ space that lead to large changes in the $\mathcal{X}$ space (see column 4 of the Figure \ref{fig:all}).

\noindent {\bf Our $w$ attack}:  Learning from the above unsuccessful approach, we minimize the distance from the original image in $\mathcal{Z}$ space by solving the optimization problem: $\minimize_{w'}\lVert h^{-1}(u',w')-z\rVert^2_2$.
This is similar to a normal adversarial minimization problem but solving the optimization problem is computationally much simpler because we are doing this on the function $h$ rather than on the much more complex function $g$. Further, the misclassification constraint is already satisfied by construction, and we are optimizing the output $w'$ instead of $z'$ directly.
The results are shown in Figure \ref{fig:all} (5th column), and the attack results are shown in Table \ref{tab: attack} (4th row). These are largely positive results for this method of generating AEs. 
However, in some cases, our attack method produces large changes in the $\mathcal{X}$ space (as seen in Figure \ref{fig:w_attacking} in the Appendix, example of "Frog").

\begin{figure}[!htbp]
    \centering
    \includegraphics[width=0.8\linewidth]{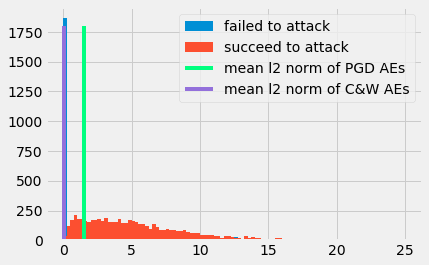}
    \caption{$L_2$ norm of adversarial examples generated by our $z$ attack}
    \label{fig:distribution}
    \vspace{-3em}
\end{figure}

\noindent {\bf Our $z$ attack}: Instead of fixing the logits to misclassify by construction, we could also generate adversarial examples by optimizing both $u'$ and $w'$: $\minimize_{u', w'}\lVert z'-z\rVert^2_2 \text{  s.t. } \argmax(h_1(z'))\neq \argmax(u)$ similar to the CW attack but just for the classifier layer $h$ instead of the full model $f$. We introduce a function as a part of loss function to encourage classification flips. i.e. $\minimize_{u', w'}\lVert z'-z\rVert^2_2 + c\cdot(-\log(2p'_t))$ , where $p'_t$ is the probability of target class. The result (Table~\ref{tab: attack}) shows comparable accuracy performance w.r.t. FGSM and PGD attacks, but it is much faster --- 6.5X faster than PGD and 915X faster than C\&W. Figure \ref{fig:distribution} shows the distribution of $\lVert x'-x \rVert_2$ among the test set. We see that the mean value of the norm is higher than for PGD and C\&W and there is a long tail. This forms the subject of our ongoing work --- how to reduce this tail. 

\paragraph{Conclusion and discussion}
In this work, we explore adversarial attacks via invertible residual networks. We design a new structure of the linear classifier and thus observe that there is a clear dependency between the modifications to the logits and the non-logits for the adversarial examples generated by today's popular AE generation techniques. 
This relationship may show a path forward to defending against AEs by observing the patterns of the non-logits. 
We also propose a new adversarial generation method from this perspective.
This method in cases generates comparable quality of AEs to PGD and C\&W, but much faster. However, the invertibility process causes the attack generation to create some AEs that are very different from legitimate examples. 
More generally, adversarial robustness of the latent representations in $\mathcal{Z}$ space may be of independent interest for training more robust classifiers.

\bibliographystyle{apalike}
\bibliography{references}\label{references}
\clearpage4
\section{Appendix}
\subsection{More details on the fully bijective network}
Householder transformation (or Householder map) is a linear transformation that represents a reflection about a hyperplane containing the origin. The hyperplane can be defined by a unit vector $u$ and the Householder map can be defined as:
$P = I - 2uu^T$ \,.
It is easy to prove that any orthogonal transformation between two vectors with the same shape could be represented as a householder transformation.
Furthermore, any linear transformation from $v\in R^n$ to $w\in R^m$, where $n>m$ could be represented as a composition of a householder transformation and a projection that only keep the first $m$ digits.
Hence, the average pooling and a full connection layer could be replaced by two Householder transformation without loss of representational power.
With only changing the last layer, our Top-1 accuracy of i-ResNets 64 achieved $92.22\%$ which is higher than the original structure $91.44\%$ on CIFAR-10.
In order to make the whole neural network bijective, we remove the zero padding introduced in the original i-ResNets code (which would make the model injective instead of bijective) and impose that the first digit of the logits $u$ ot be $0$ to guarantee bijectivity of the Softmax layer.

\begin{table}[!htbp]
\caption{Top-1 accuracy among different Lipschitz constant}
\label{tab: acc}
\centering
\resizebox{0.5\textwidth}{8mm}{
\begin{tabular}{@{}llll@{}}
\toprule
Lipschitz constant & i-ResNets & modified i-ResNets with inj\_pad & modified i-ResNets without inj\_pad \\ \midrule
0.9                & 91.44\%  & 92.22\%                         & 90.00\%                            \\
0.5                & 89.33\%  & 84.29\%                         & 83.31\%                            \\
0.1                & 76.74\%  & 72.63\%                         & 65.72\%                            \\ \bottomrule
\end{tabular}
}
\end{table}
\subsection{More Experimental Results}
\subsubsection{Performance of the modified i-ResNets}

\subsubsection{More examples}
Here's more adversarial examples generating from our methods.
\label{sec:moreexp}
\begin{figure*}[!htbp]
    \centering
    \includegraphics[width=0.95\linewidth]{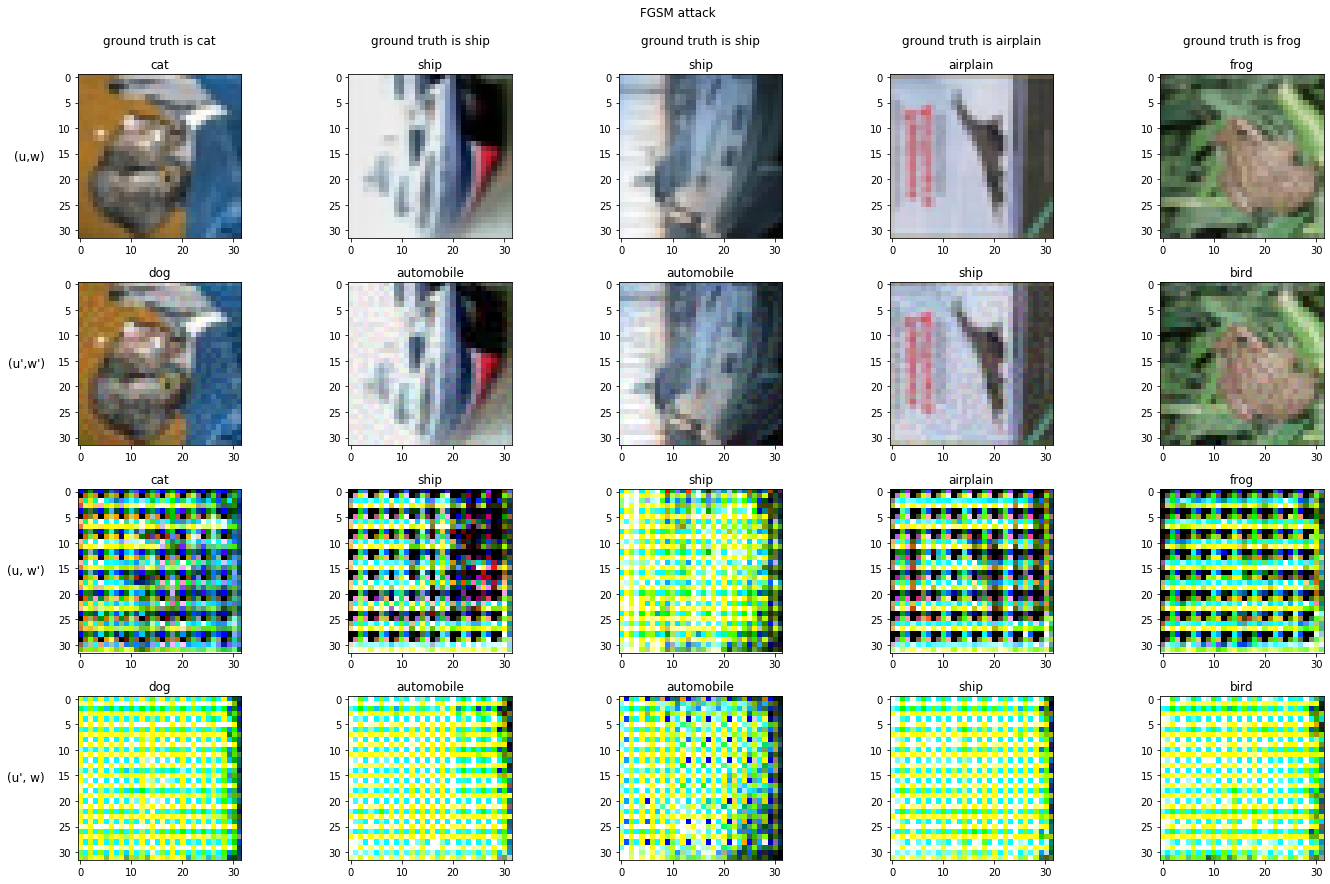}
    \caption{FGSM attack performance;  We can find that the third row and forth row are already becomes unrecognizable}
    \label{fig:fgsm}
\end{figure*}
\begin{figure*}[!htbp]
    \centering
    \includegraphics[width=0.95\linewidth]{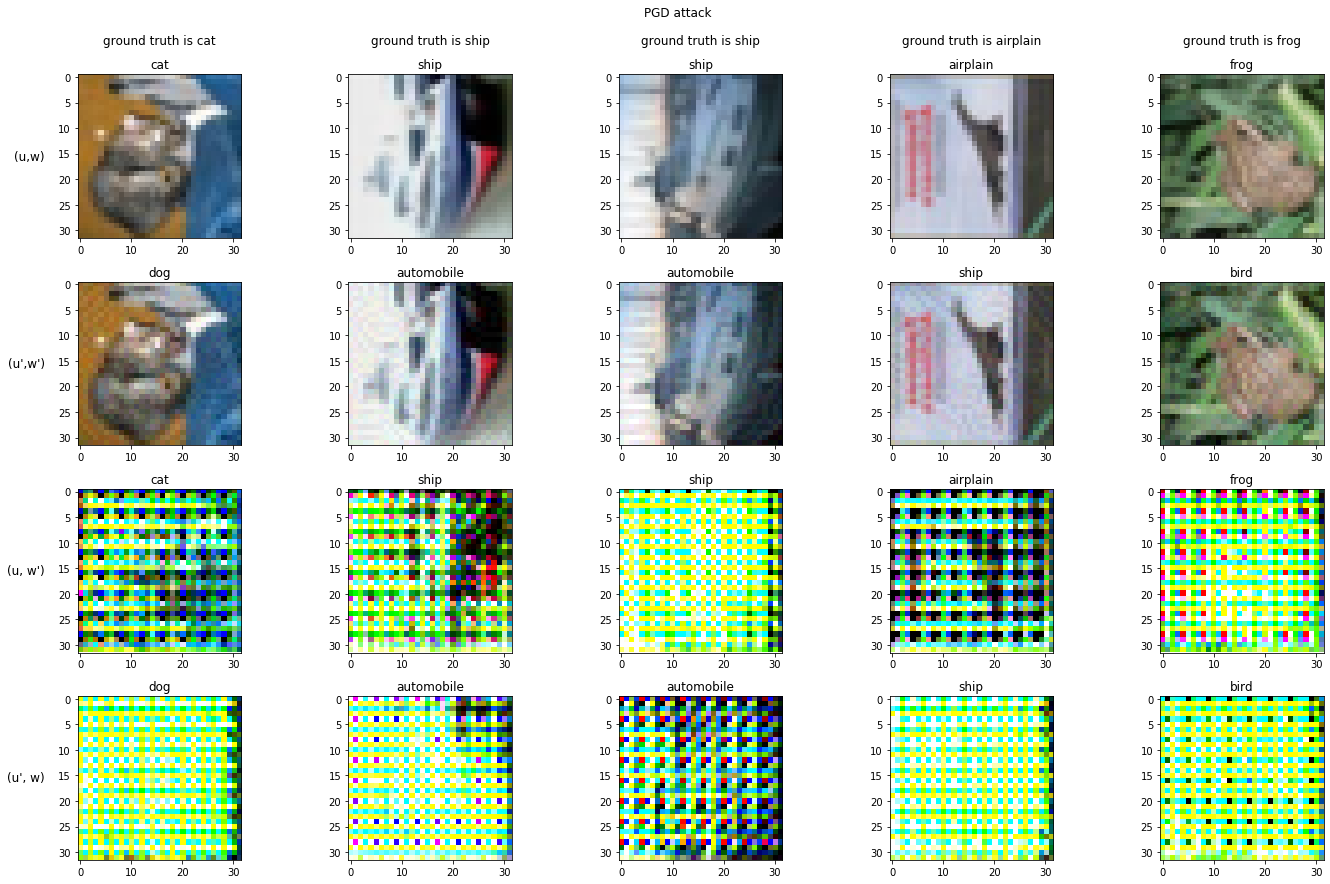}
    \caption{PGD attack performance similar results with figure \ref{fig:fgsm}}
    \label{fig:pgd}
\end{figure*}
\begin{figure*}[!htbp]
    \centering
    \includegraphics[width=0.95\linewidth]{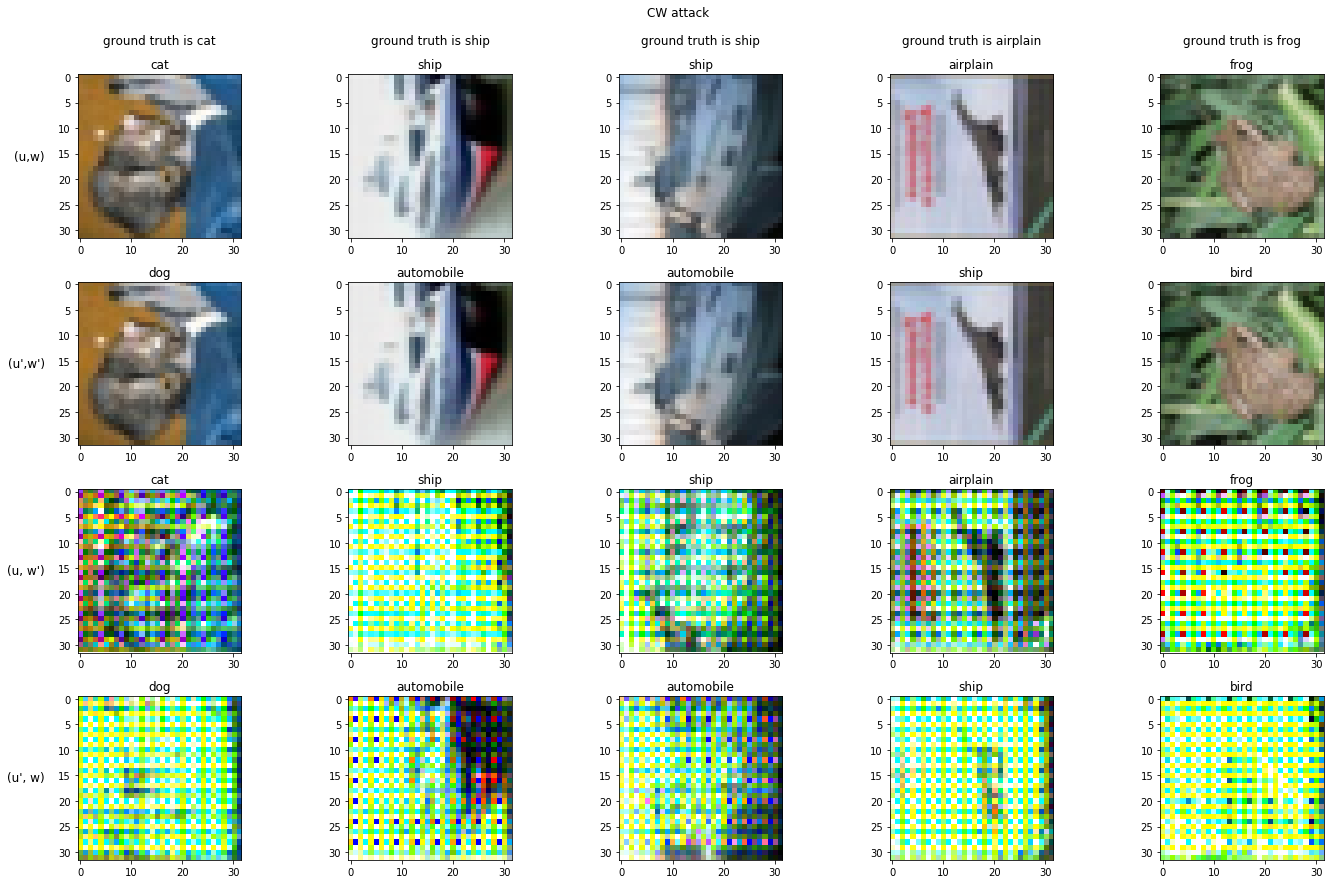}
    \caption{C\&W attack performance similar results with figure \ref{fig:fgsm} and \ref{fig:pgd} but smaller changing of $w'$ if we observe the third row}
    \label{fig:cw}
\end{figure*}
\begin{figure*}
    \centering
    \includegraphics[width=0.95\textwidth]{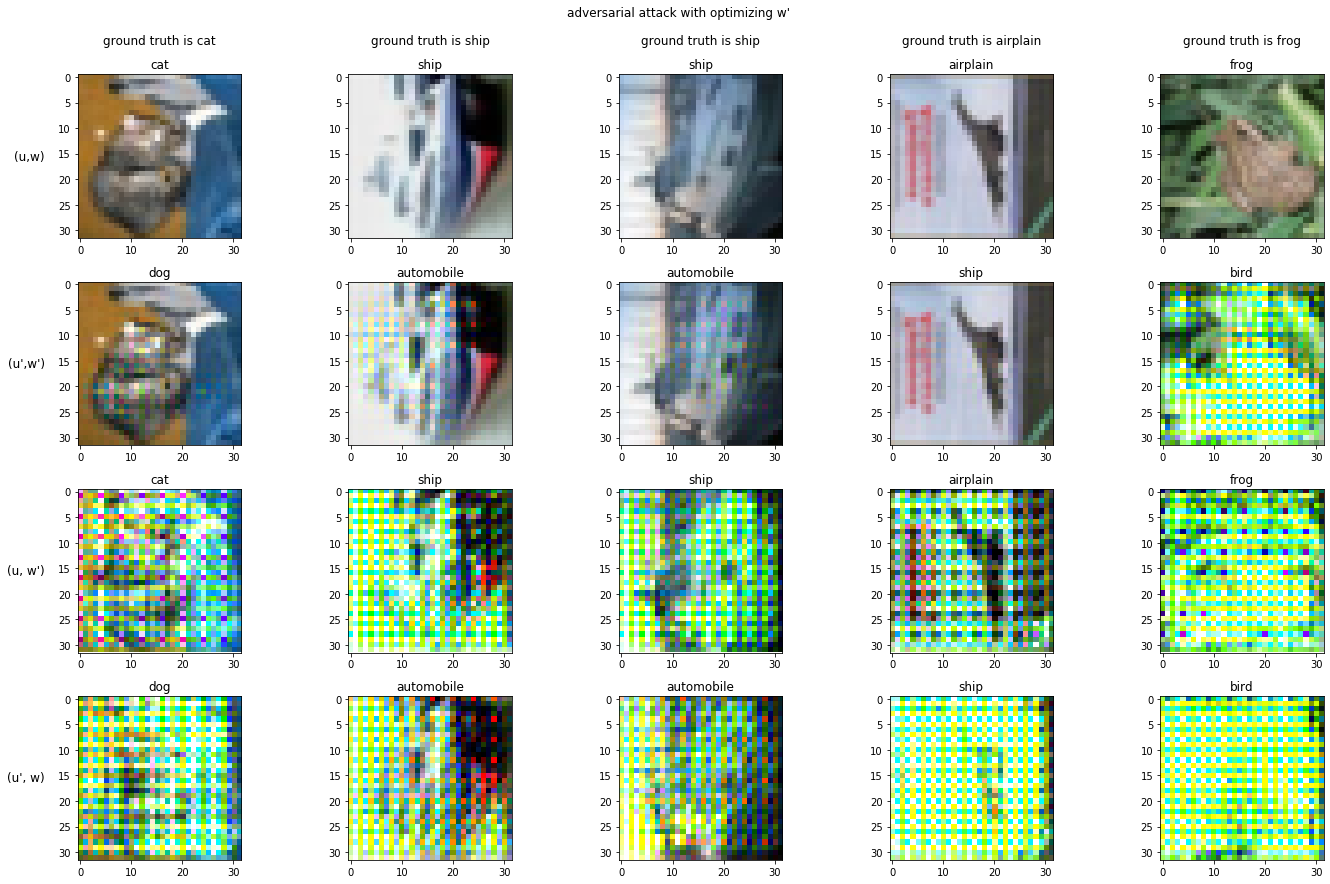}
    \caption{The first four pairs of images are almost indistinguishable to the human eye, showing the cases of successful attack. However, the fifth column shows a failure case of our attack method.}
    \label{fig:w_attacking}
\end{figure*}

\begin{figure*}
    \centering
    \includegraphics[width=0.95\textwidth]{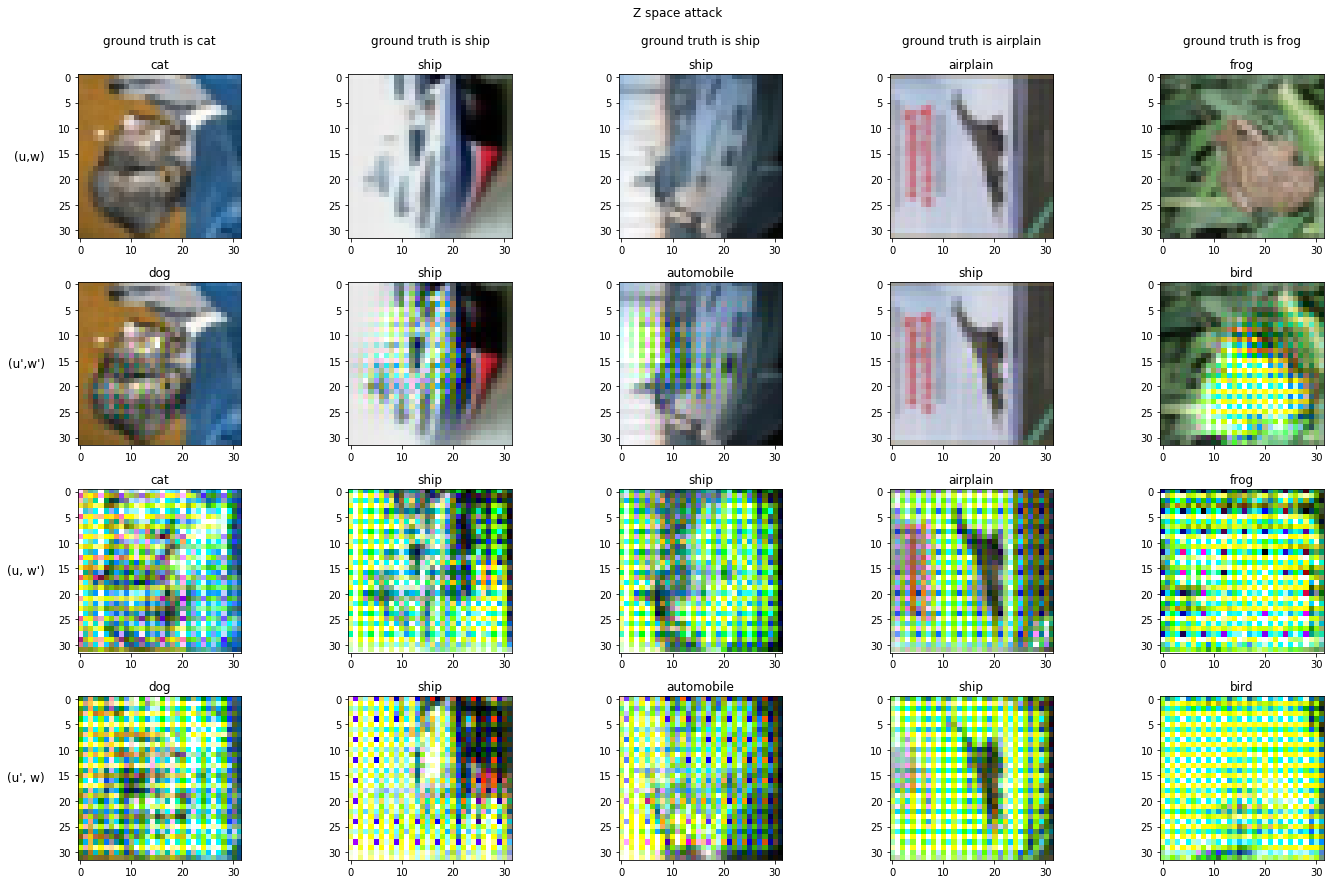}
    \caption{Attack generated from optimizing $z$. The results show similar performance to the $w$ attack.}
    \label{fig:z_attacking}
\end{figure*}
\end{document}